\documentclass[sigconf]{acmart}
\AtBeginDocument{%
  \providecommand\BibTeX{{%
    \normalfont B\kern-0.5em{\scshape i\kern-0.25em b}\kern-0.8em\TeX}}}

\setcopyright{acmcopyright}
\copyrightyear{2018}
\acmYear{2018}
\acmDOI{XXXXXXX.XXXXXXX}

\acmConference[Conference acronym 'XX]{Make sure to enter the correct
  conference title from your rights confirmation emai}{June 03--05,
  2018}{Woodstock, NY}
\acmPrice{15.00}
\acmISBN{978-1-4503-XXXX-X/18/06}

\usepackage[ruled,linesnumbered]{algorithm2e}
\usepackage{multirow}
\usepackage{subfigure}
\newtheorem{myDef}{Definition}
\begin{document}

\title{FIND: Explainable Framework for Meta-learning}

\author{Xinyue Shao}
\authornotemark[1]
\email{shaoxinyue@hit.edu.cn}
\affiliation{
  \institution{Harbin Institute of Technology}
\streetaddress{West Dazhi Street No.92}
 \city{Harbin}
  \state{Heilongjiang}
 \country{China}
}

\author{Hongzhi Wang}
\authornotemark[2]
\email{wangzh@hit.edu.cn}
\affiliation{
  \institution{Harbin Institute of Technology}
\streetaddress{West Dazhi Street No.92}
 \city{Harbin}
  \state{Heilongjiang}
 \country{China}
}

\author{Xiao Zhu}
\authornotemark[3]
\email{zhuxiao_momo@163.com}
\affiliation{
  \institution{Harbin Engineering University}
\streetaddress{Nantong Street No.145}
 \city{Harbin}
  \state{Heilongjiang}
 \country{China}
}

\author{Feng Xiong}
\authornotemark[4]
\email{xiong257246@outlook.com}
\affiliation{
  \institution{Harbin Institute of Technology}
\streetaddress{West Dazhi Street No.92}
 \city{Harbin}
  \state{Heilongjiang}
 \country{China}
}

\renewcommand{\shortauthors}{Trovato and Tobin, et al.}

\begin{abstract}
Meta-learning is used to efficiently enables automatic selection of machine learning models by combining data and prior knowledge. Since the traditional meta-learning technique lacks explainability, as well as shortcomings in terms of transparency and fairness, achieving explainability for meta-learning is crucial. This paper proposes \textbf{FIND}, an interpretable meta-learning framework that not only can explain the recommendation results of meta-learning algorithm selection, but also provide a more complete and accurate explanation of the recommendation algorithm's performance on specific datasets combined with business scenarios. This validity and correctness of this framework have been demonstrated by extensive experiments.
\end{abstract}

\ccsdesc[500]{Computer systems organization~Embedded systems}
\ccsdesc[300]{Computer systems organization~Redundancy}
\ccsdesc{Computer systems organization~Robotics}
\ccsdesc[100]{Networks~Network reliability}

\keywords{explainability,meta-learning, causal reasoning, counterfactual interpretation}


\maketitle
\section{Introduction}
The proliferation of data collection sources and the ease of data acquisition has resulted in an exponential increase in the amount of data available for analysis and decision making. Experts and scientists have proposed a number of data mining algorithms. Since different algorithms have varying inductive biases, choosing the most suited method for a given problem has been a significant challenge.

Brazdil et.al.\cite{brazdil2008metalearning} proposes to apply meta-learning to machine learning by utilizing machine learning methods as meta-algorithms to determine the mapping between a problem's meta-features and an algorithm's performance. When the user needs to know which method is optimal, the meta-algorithm employs a classification algorithm. Numerous investigations, however, have revealed that meta-learning classification algorithms' returned results are not always perfect, and when they are not optimal, the recommended results cannot convey any extra valuable information to the user due to the algorithm's black-box nature
\cite{kalousis2002algorithm}. As a result, meta-learning requires explainability.

The explainability of meta-learning is primarily concerned with two aspects of explanation. On the one hand, \emph{meta explainability} is the explainability of the meta-algorithm for algorithm selection, which summarizes the meta-knowledge underlying algorithm selection. On the other hand, \emph{recommendation explainability} is the explainability of the leaning algorithm recommended by meta-learning. The former one is about why to select the leaning algorithm, and the latter one is about the reasonability of the decision made by the selected leaning algorithm. Both these aspects bring challenges.

For meta explainability, achieving explainable meta-learning in algorithm selection is complicated by the fact that accuracy and explainability cannot coexist. From decision trees to deep learning, the penalty of enhancing the accuracy of optimal algorithm recommendation is increased algorithm model complexity, not just in terms of higher computational cost but, more crucially, in terms of increased unexplainability.

For recommendation explainability, there are several major issues to resolve. To begin, the criteria for determining what interpretation is truly required by users for various scenarios remain unclear\cite{galhotra2021explaining}. Second, the majority of existing interpretable tools\cite{friedman2017elements}\cite{goldstein2015peeking} \cite{ribeiro2016should} make the implicit assumption that data features are distributed independently and identically, oblivious to the causal relations between features, resulting in less-than-true interpretations.

In response to the first criticism on the lack of clarity in the interpretation criteria, Judea Pearl divides explanations into three levels. \cite{pearl2018book}. The first level of the theory is the association. The second level is intervention, and the third level is the counterfactual.

According to the three-level theory of interpretation, we believe that when applying the algorithm to specific problem scenarios, explanation can take place in two ways. On the one hand, it is based on intervention, in which each feature's importance for prediction is quantified by perturbing the features and evaluating the degree of change in the prediction. On the other hand, generating counterfactuals for instances is an approach for providing the user with suggestions for revising the prediction.

Numerous works have been proposed in both directions of explainability research\cite{ribeiro2016should}\cite{ribeiro2018anchors}\cite{dhurandhar2018explanations}\cite{luss2021leveraging}\cite{looveren2021interpretable}, but all of them ignore causality and make the unreasonable assumption that data features are independently and identically distributed, i.e., when one feature is modified, it has no effect on other feature values. In fact, there is a causal relation between features\cite{scholkopf2021toward}, i.e., when the value of one variable is changed, the associated feature variable is also changed.The intervention-based and counterfactual-based interpretations achieve explainability from two directions: the former is to evaluate the feature importance by perturbing them and observing the changes in predictions; the latter is to help users obtain the expected results by giving reasonable suggestions for feature modification.

Motivated by this, we attempt to adopt causal relation to the explainablity of meta-learning.
In the intervention-based interpretation, we slightly perturbed each feature and updated its latent factors in conjunction with the causal model, and finally the degree of change in the predicted outcome was used as the feature  influence score. In the counterfactual interpretation, we prioritized the features with high feature  influence score for modification, and update the latent factors according to the causal model until the desired goal isachieved.

Overall, we propose a novel interpretation framework for meta-leaning called \textbf{FIND} the framework of \textbf{F}eature \textbf{I}nfluence i\textbf{N}terpretation framework for meta-learning \textbf{D}ecision models based on causality), which assists users to choose a better model interpretation and understand the influence of each feature on the output results without going deep into the model. The contributions of this paper are as follows:
\vspace*{-0.5\baselineskip}
\begin{enumerate}
\item We propose a comprehensive framework for explicable meta-learning, \textbf{FIND}, based on causal relation. To the best of our knowledge, this is the first study for explicable meta-learning.

\item To achieve meta explainability, we develop interpretable learning algorithm recommendation approach based on the dataset's meta-features, which is able to further explore the relationship between the dataset and the algorithm while ensuring accuracy.

\item To improve recommendation explainability, we suggests a novel feature importance metric in conjunction with causality, and propose a greedy counterfactual generation approach based on our proposed explainable indicator. The generated counterfactuals are more rational as a result of our indicator's causality.

\item We evaluate the accuracy and validity of each module of the \textbf{FIND} framework in real datasets, and the superiority of our method is fully demonstrated in comparison with existing methods in various fields.
\vspace*{-0.5\baselineskip}
\end{enumerate}

\section{Related work}

The term meta-learning originated in psychology, and afterwards \cite{brazdil2008metalearning} advocated applying it to machine learning. The STATLOG project recommended suitable machine learning algorithms through decision trees. Gama\cite{gama1995characterization} proposed the use of regression algorithms to forecast the performance of learning algorithms, employing three different regression approaches to estimate the error. Bradzil\cite{brazdil2003ranking} were the first to construct and evaluate machine learning algorithm rankings using the k-NN algorithm.

The advantage of employing classification algorithms as meta-algorithms is that they have a vast choice of algorithms, but the output of classification algorithms is a single optimal algorithm. When the selected algorithm is not optimal, the disadvantage of classification algorithm as a meta-algorithm is that it does not give the user with alternative algorithm information. For these reasons, this paper argues that it is necessary to conduct interpretable research on the meta-learning process,  which is still in its infancy, and the interpretable work proposed by scholars is still focused on a single decision model for a specific problem.

Several surveys \cite{chakraborty2017interpretability} \cite{zhang2018visual} \cite{gilpin2018explaining} \cite{adadi2018peeking} \cite{guidotti2018survey} \cite{du2019techniques} \cite{fan2021interpretability} have already helped us summarize a list of the numerous interpretable approaches that have been proposed in recent years. Works on the explainability of decision methods can be broadly classified into model-specific and model-agnostic approaches depending on whether they are restricted to a specific model class.

We will concentrate on these methods in this work since they are usually model-agnostic, have the advantage of isolating the interpretation from the method \cite{ribeiro2016model}, and exhibit more flexibility, allowing users to utilize them in conjunction with whatever method they are interested in as needed.

Visualization, feature attribution, proxy models, and counterfactual explanations are representative model-agnostic interpretation methods. In the remaining part, the relevant work of these methodologies will be discussed.

In terms of visualization, Matthew D. Zeiler et al. \cite{zeiler2014visualizing} proposed employing deconvolution to transfer hidden layer features back to pixel space and visualize them. By estimating the model gradient, Jost Tobias et al. \cite{springenberg2014striving} suggested a guided-backpropagation strategy for visualizing high-level properties. A method for extracting the image-specific class saliency map from the image has been proposed by Simonyan et al. \cite{simonyan2013deep}. Sundararajan et al. \cite{sundararajan2017ax} attributed deep network prediction performance to input features. Two essential attribution axioms are first identified: sensitivity and implementation invariance, and then a new attribution method called "Integrated Gradients" is devised based on these two axioms. Zhou et al. \cite{zhou2016learning} proposed CAM (class activation map), which is a method for linearly weighting the feature map of the middle layer to recognize the discriminative area in an image. To improve the convenience of CAM, Selveraju et al. \cite{selvaraju2017grad} and Chattopadhyay, A., et al. \cite{chattopadhay2018grad} proposed Grad-CAM and Grad-CAM++, respectively. In addition, Wang et al. \cite{wang2020score} found that the visualization results generated by the previous gradient-based CAM method were not visually clean enough, so they proposed a visually interpretable confidence score-based method, called Score-CAM.

Feature attribution mainly refers to ranking or measuring input features and quantifying the impact of each feature on decision-making results. JH Friedman et al. \cite{friedman2017elements} presented PDP, which calculates the average marginal effect of a given feature on the expected outcome using partial dependence functions.

In order to solve the problem that PDP graphs cannot show heterogeneous effects, Goldstein et al. \cite{goldstein2015peeking} proposed ICE, which can reveal heterogeneous relationships created by interactions.  Daniel W Apley \cite{apley2020visualizing} directly faced the feature correlation problem and proposed the ALE, which can remain valid even when feature correlation exists. Lundberg and Lee \cite{lundberg2017unified} proposed the SHAP method based on the game-theoretic theory of Shapley \cite{nowak1994shapley} values to calculate the marginal benefit contribution of features.

Training proxy models is another method for achieving model-agnostic explainability. By combining input data with the same average hidden unit activation levels, Setiono et al. \cite{setiono1995understanding} derived hidden layer rules for neural networks. Saad \cite{saad2007neural} and Thrun \cite{thrun1995extracting} employed a pedagogical technique to extract rules with similar input-output correlations. The LIME model proposed by Marco Tulio Ribeiro \cite{ribeiro2016should}  It establishes a local interpretable linear model to approximate the local black-box methods' predictions, with the linear model's coefficients indicating the contribution of the features. Ribeiro \cite{ribeiro2018anchors} provided anchors based on LIME to explain the model's local behavior in terms of rule sets, allowing the user to infer model behavior from the rules.

Counterfactual interpretation is a general explainability paradigm that assists users in determining how to alter decision outcomes with minimal changes to the original features. On structured data and colorful image data with complicated structures, Dhurandhar et al. developed counterfactual generation methods \cite{dhurandhar2018explanations} \cite{luss2021leveraging}. \cite{yang2021model} established a counterfactual interpretation framework using CGAN and training with umbrella sampling. Counterfactual generation is characterized in \cite{looveren2021interpretable} as an optimization issue with the goal of minimizing the loss function in order to discover the data point that conforms to constraints and is the closest to the corresponding instance. \cite{le2020grace} developed a new framework, GRACE, that discovers key features of samples, generates comparative samples based on these features.

This paper is the first interpretable research in the field of meta-learning and clarifies that the explainability of meta-learning includes meta explainability and recommendation explainability. In meta explainability, a series of experiences that can help analysts to select models are summarized; in recommendation explainability, causality is fully combined to achieve feature influence computation and counterfactual generation.


\section{Framework}
\begin{figure}[h]
  \centering
  \includegraphics[width=\linewidth]{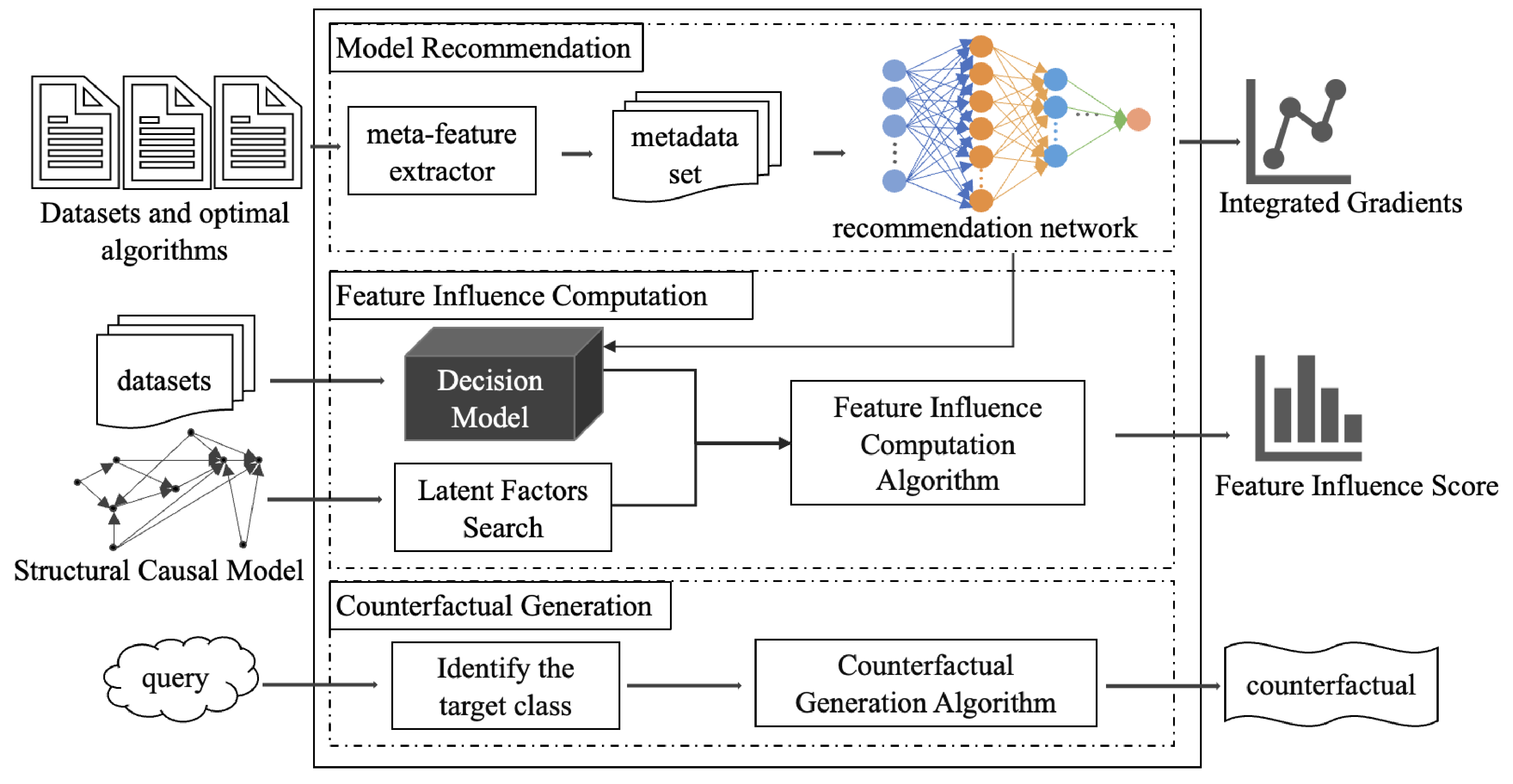}
   \caption{The workflow of \textbf{FIND} }
   \label{workflow}
\vspace*{-0.5cm}
 \end{figure}

In this section, we introduce the proposed \textbf{FIND} framework. We divide the task into three parts: model recommendation, feature influence  computation, and counterfactual generation. For the model recommendation section, we compute the integrated gradient of features for the meta-learning, which can reveal the degree of influence of each feature on the results and thus achieve meta explainability. Feature influence computation and counterfactual generation achieve recommendation explainability by quantifying feature importance and providing feature modification suggestions to change the prediction results, respectively. The workflow is shown in  Figure~\ref{workflow}.



The first component is model recommendation.
This component takes common machine learning datasets
and the best performing models
we tested on each dataset as the training set. This paper constructs a meta-feature extractor to retrieve the meta-features describing the dataset, which combined with the optimal algorithm constitute the metadata set. For this metadata set, a recommendation network is constructed, which is trained to recommend suitable algorithms for the dataset, and by calculating the integrated gradient. a series of experiences that can help analysts to select models can be inducted. 
This component will be described in detail in Section 4.

The second component is feature influence computation. In response to the limitation that existing feature attribution methods are based on the independent homogeneous distribution of data features, we propose a method that combines causality to determine the importance of each feature for this algorithm. The input has two parts. The first one is the dataset that needs an appropriate model. 
The second one is the causal structure model, which is a formal expression of causality.
To integrate causality for a more precise interpretation, the latent factors for each feature in the causal structure model need to be discovered using the latent factors search. Then the feature influence computation allows us to quantify the influence of features on the prediction.
This component will be described in detail in Section 5.

The third component is counterfactual generation, which proposes a greedy counterfactual generation algorithm that satisfy the casual relations between features.
Since this is an instance-based method, a specific instance is needed as query,
and immediately after that,
the target class of counterfactual generation needs to be specified. The final output of the method is a suggestion that helps the query to cross the decision boundary and be predicted as a target class. 
This part will be described in detail in Section 6.

\vspace*{-0.2cm}

\section{Model Recommendation}
Model recommendation aims to to select the most appropriate algorithm from the many options available for the given situation.


The problem is defined as follows. $P$ stands for problem space. Extract the measurable features of a given problem instance $x \in P$ and indicate them as $f(x) \in F$. The algorithm space $A$ denotes the set of methods, including $\{S_1, S_2, \cdots ,S_m \}$, and the performance space $Y$ is the result of each method's performance. The objective of model recommendation is finding $\alpha$ that satisfies constraint $\alpha=\mathop{\arg\max}\limits_{S_i}M(S_i(f(x)))$.


The simplest way is to test all available algorithms on this dataset, and then pick the top performing method. But it requires a lot of computational resources and time. The other way is guided by expert empirical knowledge, but it is not always dependable and unscalable.

To tackle this problem, we applied meta-learning in model recommendation. Dataset statistics are retrieved as meta-features, candidate algorithms are tested on the datasets, and then a meta-algorithm is applied to the metadata to generate a mapping relationship between the meta-features and algorithm performance. Finally, the integrated gradient is calculated for the meta-algorithm in order to achieve meta explainability. Because based on the gradient values, we are able to generalize the relationship between some dataset meta-features and model recommendation results.

Meta-feature is defined as follows.

\vspace*{-0.5\baselineskip}
\begin{myDef}
  \label{Meta-feature}
  (Meta-feature). Let $D$ denote the set of all datasets, and $A$ signify the set of selected statistical algorithms. For each dataset $d_i \in D$, the meta-feature is that the feature extracted by the algorithm $a_j \in A$ , which is denoted as $m_{ij}=a_j(m_i)$. $M_i=\{m_{i1},\cdots,m_{i|A|} \} $ is the meta-feature description for $d_i \in D$.
\vspace*{-0.5\baselineskip}
\end{myDef}

The extraction of high-quality meta-features from the dataset is a necessary condition for accurate and successful algorithmic recommendation.
Three types of meta-features are selected, including simple features, statistical features and information features. The simple features indicate the basic structure, such as its size, the number of attributes, and the number of categories. The statistical features, such as geometric mean and variance, primarily indicate the central tendency of the number and the dispersion of the attributes. The information features reflect the degree of correlation between different attributes and the consistency of the data, such as the information entropy of the attribute variables.

\begin{figure}[h]
  \centering
  \includegraphics[width=\linewidth]{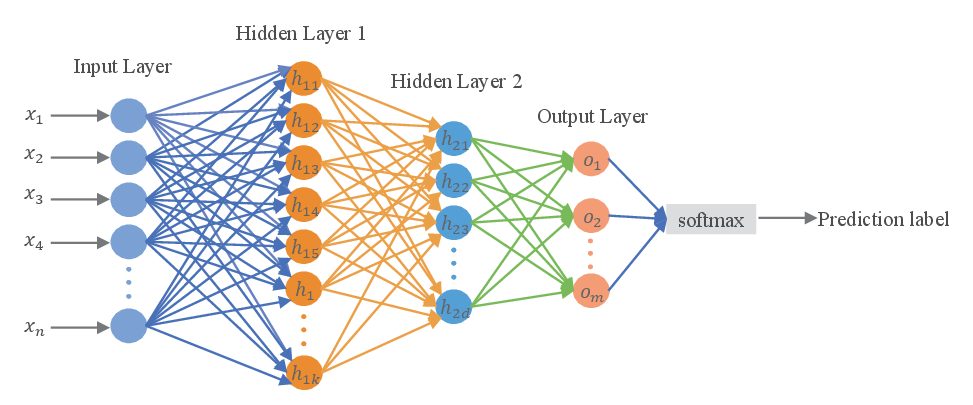}
   \caption{Recommendation Network}
   \label{classifier}
\vspace*{-0.5cm}
\end{figure}

To increase the accuracy of the ideal algorithm recommended, we employ a neural network as a meta-learning method.
The classifier mainly consists of one input layer, two hidden layers, and one output layer.  A simple mapping is added to select the category with the highest probability as the output label to make the classification result more than just the probability vector of each category. The whole network is illustrated in Figure~\ref{classifier}. For the sake of simplicity, the activation function after the hidden layer and the dropout layer are not represented in the figure; however, each hidden layer has a fully connected operation, an activation function, and a dropout.

\vspace*{-0.5cm}

If a total of $n$ meta-features are extracted for each dataset $d_i \in D$, the meta-feature description is $M_i=\{ m_1,\cdots,m_n \}$. The weights $W$ and bias $B$ between the layers are initialized according to the parameters, and each meta-feature are normalized and concatenated to generate $X_i=\{ x_1,\cdots,x_n \}$ as the input.

If there are $k$ neurons in the first hidden layer and $d$ neurons in the second hidden layer, the final output of the first hidden layer is $H_1=\{h_{11},\cdots,h_{1n} \}$, and the output of the second layer is $H_2=\{h_{21},\cdots,h_{2d} \}$. $H1$ can be calculated according to input $X_i$, connection weight parameters $W_i \in \mathbb{R}^{n \times k}$, and bias $B_1 \in \mathbb{R}^{1 \times k}$ between the input and the first hidden layers.

\vspace*{-0.5\baselineskip}
\begin{equation}
  H_1=\psi(W_1X_i+B_1)~~~~i=1,2,\cdots,t
\vspace*{-0.5\baselineskip}
\end{equation}
where $\psi(\cdot)$ denotes the activation function, and $t$ denotes there are $t$ instances in the training set.

More specifically, for $h_{1j} \in H_1$, this formula can also be written in detail as follows, where $w_{1ij} \in W_1$ and $b_{1j} \in B_1$.
\vspace*{-0.5\baselineskip}
\begin{equation}
  h_{1j}=\psi \left( \sum_{i=1}^n {w_{1ij}} \times x_{i}+b_{1j} \right)~~~~
  j=1,2,\cdots,k
\vspace*{-0.5\baselineskip}
\end{equation}

In order to further enhance the nonlinear representation of the network, a nonlinear function is introduced as the activation function. The $ReLU$ activation function is adopted in the recommendation network. Its computation cost is substantially lower than that of $sigmoid$ and $tanh$ functions, which need derivation and division, and it solves the problem of gradient disappearance in $sigmoid$ function backpropagation. It significantly reduces the occurrence of overfitting by making a portion of the neuron's output zero. For a given element $x$, the $ReLU$ activation function is defined as:

\vspace*{-0.5\baselineskip}
\begin{equation}
  \psi(x)=max(x,0)
 \vspace*{-0.5\baselineskip}
\end{equation}

The connections between the two hidden layers, as well as the connection between the second layer and the output layer, are similar to those described previously.

The output layer's result is not an obvious description of the likelihood that $X$ belongs to each class, so a $softmax$ function is required to convert the output values to a probability distribution.
\vspace*{-0.5\baselineskip}
\begin{equation}
  \hat{y_i}=softmax(o_i)=\frac{e^{o_i}}{\sum_{j=1}^m{e^{o_i}}}  ~~~~
  i=1,2,\cdots,m
\vspace*{-0.5\baselineskip}
\end{equation}

where $\hat{y_i}$ denotes the probability that $X_i$ is predicted as class $i$.

Creating a label probability vector $Y_i=\{y_1,\cdots,y_m\}$ for the true value of the sample $X_i$, with $y_i =1$, $i$ is the valid class of $X_i$, and the remaining values equal to $0$. Finally, the cross-entropy loss function is used to compute the difference between the prediction probability and label probability.

\vspace*{-0.5\baselineskip}
\begin{equation}
  \mathcal{L}=\frac{1}{t}\sum_{i=1}^t{L_i}=-\frac{1}{t}\sum_{i=1}^t{\sum_{c=1}^m{y_{ic}\log(\hat{y_{ic}})}} ~~~~
\vspace*{-0.5\baselineskip}
\end{equation}

where $t$ denotes the number of instances in the training set, $m$ denotes the number of classes, and $y_{ic}$ denotes the prediction probability that the observed sample $X_i$ belongs to class $c$.

After iterative training of forward propagation, computation of loss values, backward propagation, and gradient update, the recommendation network is able to recommend appropriate decision methods based on the meta description of a problem.

The network is black-box and users cannot intuitively understand the correlation between meta-features and recommended models. In order achieve the explainability of meta-learning, we attempt to understand the influence of each meta feature on the recommendation result by calculating its gradient in the network.

The direct gradient is based on a Taylor expansion that uses the $i^{th}$ component of the gradient vector as the $i^{th}$ feature's importance indicator. It can explain certain prediction outcomes in many circumstances, but it has an obvious drawback, i.e., the gradient tends to $0$ when the input feature value is at the gradient saturation stage, which is often the negative half-axis of the $ReLU$ function. Nevertheless, this does not mean that the feature is worthless. As a result, the direct gradient's sensitivity is insufficient as a good indicator of importance.

To overcome the shortcomings of direct gradient, we adapt an interpretable technique called integrated gradient~\cite{sundararajan2017ax}. This is a strategy that combines direct gradient and back-propagation based attribution and adheres to the sensitivity and realization invariance assumptions. By calculating the integrated gradient of the recommendation network, we are able to understand the influence of individual features on the recommendation results. The integrated gradient is calculated as follows.

Let the input be $x$, the baseline value be $x'$, and the function mapping be $F$. The integrated gradient for the $i^{th}$ dimension of the input can be expressed as:
\vspace*{-0.5\baselineskip}
\begin{equation}
  IG_i=\left(x_i - x'_i\right) \times \int_{0}^{1} \frac{\partial F(x'_i+\alpha\times (x_i - x'_i))}{\partial x'_i} \, d\alpha
\vspace*{-0.5\baselineskip}
\end{equation}

It is clear that from the above equation that the integrated gradient algorithm only considers the model's input and output without considering the model's internal. Since the function is differentiable everywhere, the integrated gradient satisfies the implementation invariance. The integrated gradient takes into account the gradients of all points along the path by selecting an infinite number of integral points between the baseline and the input value for integration and summation, so it is no longer limited by the gradient of $0$ at a certain point, solving the gradient saturation problem and satisfying the sensitivity.

The detailed workflow of this component is shown in Figure~\ref{workflow}. The network automatically recommends an appropriate decision model based on the meta description of the problem and is interpretable in conjunction with the integrated gradient, which explicitly provides the user with a direct relationship between the problem description and the selection of decision model. Thus, the proposed technique enables the explainability of meta-learning in the algorithm selection process.
\section{Feature Influence Computation}
Meta-learning explainability also includes the explainability of the model recommended by meta-learning for a specific problem. Analyzing the impact of features on predictions is one of the major types of approaches to achieve model explainability \cite{gilpin2018explaining}
, referred to as feature attribution, but existing feature attribution methods
primarily calculate the direct effect of a change in the feature on the output, regardless of the causal relations between the features. The existence of causal relations between features indicates that changes in some features cause changes in other features, and that changes in outcomes are the consequence of the combined influence of these two types of features. Therefore, the calculation of the feature importance without causality is incomplete and detrimental to explainability.
To more precisely quantify the value of each feature in the model, we propose a method for calculating feature influence that incorporates causality.

We have repeatedly emphasized that explainability cannot be separated from causality because causality exists in many fields, including economics, law, medicine, and physics, although it is difficult to describe. In layman's words, causality exists when one event causes the occurrence of another, and the latter is the result of the former. It is important to note that causation differs from correlation. The correlation is symmetric, and the correlation between the random variables $A$ and $B$ can be defined as:
\vspace*{-0.3\baselineskip}
\begin{equation}
corr(A,B)=\frac{Cov(A,B)}{\sqrt{Var(A)Var(B)}}
\vspace*{-0.3\baselineskip}
\end{equation}
Obviously $corr(A,B)=corr(B,A)$, so the symmetry of the correlation is proved.

Causality is asymmetric, which means that event $A$ causes event $B$ does not indicate that event $B$ triggers event $A$.

Experts from numerous domains have created various representations of causality models, such as causality diagrams, structural equations, logical statements, and so on, to convey causality vividly
. Judea \cite{pearl2018book} pointed out that the structural causal model (SCM) is the clearest and most understandable of them all, defined as follows.

\vspace*{-0.5\baselineskip}
\begin{myDef}
  \label{Structural Causal Model}
  (Structural Causal Model). The structural causal model is defined as the ordered triplet $<U,V,f>$, in which $U$ denotes a set of exogenous variables, $V$ denotes a set of endogenous variables, and $f$ denotes equations that indicates the relationships between variables.
\vspace*{-0.5\baselineskip}
\end{myDef}

Since SCM is a generalization of Bayesian networks\cite{pearl2018book}, it is represented by a directed acyclic graph(DAG) $G$. The edges of $G$ are the relations described by functions in $f$, and the nodes are the variables in $U$ and $V$. If $X=f(Y)$, a directed edge in $G$ will point from $X$ to $Y$, indicating that $X$ is the parent node of $Y$.

This helps find the latent factors for each feature.In order to measure the contribution of each feature to the model more comprehensively, we propose a feature influence  computation method that combines latent factors, which can integrate the direct and indirect effects of features on the output, and thus achieve the explainability of the decision method.

The potential influences on the features can be expressed as descendant nodes in the SCM. Therefore, determining the latent factors can be formulated as a search problem for the descendants in a directed graph, as shown in Algorithm~\ref{firstalg}. Throughout the search process, the algorithm combines the qualities of the causal graph itself to develop a pruning strategy. Since not all features are causally associated in the actual problem scenario, the latent factors of those features that do not appear in the causal graph are denoted as the $\varnothing$. Furthermore, since exogenous nodes cannot be descendants of any node, the search for child nodes is only performed in $V$
If all of the nodes in $V$ are already visited in the latent factors, the search is terminated and the latent factors are output.

\begin{algorithm}\scriptsize
\caption{Latent Factors Search}\label{firstalg}
\KwIn{Graphical SCM $G$, Initial feature $X$, Endogenous variables $V$, Exogenous variables $U$}
\KwOut{Latent factors $Latent$}
\If{$X \notin (U \cup V)$}{
\Return $\varnothing$\;
}
Initialize the search queue $Queue \leftarrow Queue\cup X$\;
\While{$Queue \ne \varnothing$ and $V \ne \varnothing$}{
$t \leftarrow Queue.top()$\;
childlist = getChildNode($t,G,V$)\;
\For {$d \in childlist$}{
$Latent \leftarrow Latent \cup d$\;
$Queue \leftarrow Queue \cup d$\;
$V \leftarrow V \setminus d$\;
}
}
\Return $Latent$
\end{algorithm}

Although the output latent factors lack a sequential relationship, we require a sequence to determine which factors should be updated first. As a result, it is critical to generate a linear sequence for the causal graph's vertices. As previously stated, the causal graph is a directed acyclic graph,  which indicates there must be at least one topological sorting.


When computing the value of any latent factor affected, the topological order of this causal graph is merged in the computation of feature influence to ensure that other features which potentially affect the factor have been updated.


After determining the update order, we need to determine how to measure the impact of the features. In this paper, we use the influence of feature changes on the predicted results is used as the index to measure the feature importance. A sufficiently small perturbation $\Delta{x_i}$ is added to the given target feature $x_i$ , whose feature influence value $S_i$ is 
, which is defined as follows.

\vspace*{-0.3\baselineskip}
\begin{equation}
  S_i=\frac{\hat{y}-y}{\Delta{x_i}}=\frac{M(X')-M(X)}{\Delta{x_i}}
\vspace*{-0.3\baselineskip}
\end{equation}

where $M$ is the adopted decision method, $X$ is the original input, $y$ is the output of the decision method without perturbing the features, $X'$ is the input of the features after updating the latent factors affected by $x_i$ based on causality, and $\hat{y}$ is the result after perturbing $x_i$ and updating the latent factors. The whole calculation process is shown in Algorithm \ref{secondalg}. In this way the influence of each feature on the model prediction can be quantified in conjunction with causality, thus achieving the goal of explainability.

\begin{algorithm}\scriptsize
\caption{Feature Influence Calculation}\label{secondalg}
\KwIn{Feature representation $X=\{ x_0,\cdots,x_i,\cdots,x_n\}$, Initial feature $x_i$, Decision method $M$, Graphical SCM $G$}
\KwOut{Feature Influence $S_i$}
$y=M(X)$ \tcp*[h]{the original output}\;
$Latent_i=Latent Factors Search(G,x_i,V,U)$\;
\While(\tcp*[h]{Get Topological Sorting of $G$}){node set $\ne \varnothing$}{
Add the nodes with degree 0 to $Order$\;
Delete these nodes and all edges from them\;
}
$x'_i=x_i+\Delta x_i$\;\tcp*[h]{add perturbations to target feature}\;
\For(\tcp*[h]{execute in topological order}){$node:Order$}{
\If{$node \in Latent$}{
$node=update(node)$\tcp*[h]{for each node, construct a predictor based on the causal graph}\;
}
}
$  S_i=\frac{\hat{y}-y}{\Delta{x_i}}=\frac{M(X')-M(X)}{\Delta{x_i}}$\;
\Return{$S_i$}
\end{algorithm}

\section{Counterfactual generation}
If feature importance is to assess the sensitivity of the decision method to each feature in a positive way to achieve explainability. Then counterfactual is to provide explanations in a desired result-oriented manner by modifying the features as little as possible to achieve the target result.
Quantifying the impact of features on the model alone provides limited explanatory power, and in many cases, what users want most is to  know how to change the predicted outcome through effort.

For instance, if one requests a loan at a bank, the bank will utilize the black-box decision method to assess the applicant's creditworthiness and decide whether to grant her the loan. If the applicant's application is refused as a result of the decision method, the applicant will be left wondering why her application was rejected. The conventional explanation technique can help the applicant which factors are critical in the decision, whereas the counterfactual explanation can teach the applicant how to alter her conduct to obtain a loan authorized.

Counterfactual reasoning is a fundamental way of thought in human awareness\cite{chiappa2019path}
, in which people frequently consider whether occurrences will alter by supposing that some aspects of the event have changed. The counterfactual interpretation in the black-box model provides users with feedback on which they can act to achieve the desired outcome in the future.

We define the counterfactual interpretation based on comparable similarity introduced by Lewis~\cite{lewis1974causation} as follows.


\vspace*{-0.5\baselineskip}
\begin{myDef}
  \label{counterfactual interpretation}
  (Counterfactual Interpretation) Given a classification model $f_\theta:\mathbb{R}^d$
  \vspace*{-0.2\baselineskip}
  \begin{equation}
    x^{cf}=\mathop{\arg\min}\limits_{x^*\in C}l(x^*,x) \mathrm{s.t.} f_\theta(x)=t, f_\theta(x^*)=t'
\vspace*{-0.5\baselineskip}
  \end{equation}
 \vspace*{-0.5\baselineskip}
\end{myDef}

where $x^{cf}$ is the counterfactual interpretation for instance $x$, $C$ is the counterfactual universe of the observed data space,  $l$ is the distance metric between $x^*$ and $x$. Although $t,t'\in \mathbb{R}^d$ are the predicted outputs of $f_\theta$ for $x$ and $x^*$, respectively,  $t' \neq t$ denotes counterfactual interpretations cause model decisions to be reversed.

As indicated in the definition, the essence of counterfactual interpretation is that a counterfactual is a sample of data from some distribution that may be used to reverse model judgments as needed while remaining similar to the query input.

The primary difficulty that counterfactual generation faces at the moment is that the generated counterfactuals are not realistic, which results from neglecting causality and the value domain limits imposed by the feature values themselves.

Therefore, we design a counterfactual generating algorithm based on feature influence score, as indicated in Algorithm \ref{thirdalg}.  For a given instance $x$, the algorithm applies a greedy strategy to discover the counterfactual interpretation $x^{cf}$ that is as near to $x$ as possible.

First, the influence score of each feature in $x$ is calculated and denoted as $S$. Then, the feature $x_k$ with the largest influence score $S_k \in S$ is selected for modification.

\vspace*{-0.5\baselineskip}
\begin{equation}
x_k=x_k-\alpha \times S_k
\vspace*{-0.5\baselineskip}
\end{equation}

The largest value signifies that the classification result is most sensitive to the relevant feature perturbation. If one intends to fulfill the purpose of modifying the decision result with the least cost, one intuition is to select for the feature that has the most impact on the result. Second, considering the causal relationships between features, the latent factors influenced by $x_k$ need to be changed. $P$ is a projection that insures that the final $x^{cf}$ seems more realistic, and the defining the domain space dom($X$) contains the maximum, minimum, data type, etc. The complete search operation pauses when the decision result is revised.

\begin{algorithm}\scriptsize
\caption{Counterfactual Interpretation Generation}\label{thirdalg}
\KwIn{Decision method $M$, Query instance $X=\{x_0,\cdots,x_n\}$, Learning rate $\alpha$}
\KwOut{Counterfactual Interpretation $x^{cf}$}
Initialize $x^{cf} \leftarrow x$\;
$t\leftarrow M(x)$\;
\While{$M(x^{cf})=t$}{
compute all feature influence score of $X$, and denote as $S$\;
choose the $S_k=max\{S_0,\cdots,S_n\}$\;
$x_k=x_k-\alpha \times S_k $ \;
$x^{cf} \leftarrow update Latent_k$\tcp*[h]{update latent factors based on causality,the process is shown in Algorithm \ref{secondalg}}\;
$x^{cf}\leftarrow P(x^{cf},dom(X))$\;
}
\Return{$x^{cf}$}
\end{algorithm}

In the process of generating counterfactuals, we fully take into account the causal relations between features and the actual taking constraints of feature values, 
and are thus able to generate more plausible counterfactuals to achieve counterfactual interpretation of meta-learning recommendation models in specific problems.

\section{Experiments}
In this section, we conducted extensive experiments to verify the efficiency of each module in the FIND framework.
\subsection{Evaluation on model recommendation}
In this section, we will evaluate the module's accuracy in recommending algorithms for various problems and the method's performance on specific problems.

\subsubsection{Experimental settings}
We created a metadata set for the dataset by collecting data of target attribute categories, the number of category attributes, the number of numeral attributes, the information entropy of attribute values within each category, and other factors. Each dataset has 23 meta-features, whose label corresponds to the classification method that performed the best on that dataset throughout the test. During the training process, we set the dropout rate to 0.5 and the learning rate to 0.001 for each layer.

\subsubsection{Accuracy and validity}
To evaluate the accuracy of the recommendation module's recommended decision algorithm, we compare its accuracy to that of Random Forest and XGBoost, as shown in Table \ref{tab:acc}. As can be seen, this module's accuracy is higher than $80\%$, which is significantly greater than the accuracy of the other two methods. To further validate whether the recommended methods are still effective in specific problems when the recommended methods are inconsistent with the labels in the metadata set, a subset of the dataset with inconsistent predictions and labels is selected. We measure the $precision$, $recall$, and $F-measure$ score of the prediction methods and labeled methods on each dataset of the subset.

\begin{table}
\vspace*{-0.5cm}
  \caption{Accuracy of recommended methods}
  \label{tab:acc}
  \begin{tabular}{cccc}
    \toprule
    Methods & Random Forest& XGBoost & Our Module\\
    \midrule
    Accuracy & 61.61\% & 75.00\% & 86.60\%\\
    \bottomrule
  \end{tabular}
 \vspace*{-0.5cm}
\end{table}

\begin{table*}[!htbp]\scriptsize
\vspace*{-0.5cm}
\centering
\caption{Performance comparison of recommended and labeled methods}
\label{tab:per}
\begin{tabular}{c|ccc|ccc|ccc}
\hline
\multirow{2}{*}{Dataset} & \multicolumn{3}{c|}{Label Model} & \multicolumn{3}{c|}{Recommended Model}& \multicolumn{3}{c}{Other Metrics}\\
\cline{2-10}
& Precision & Recall & $F-measure$  & Precision & Recall & $F-measure$ & $\mathcal{F}^1$ & $\mathcal{F}^2$ & $\mathcal{F}^3$ \\
\hline
anneal & 0.993 & 0.993 & 0.993 & 0.982 & 0.981 & 0.981 & 0.989 & 0.988 &	0.988\\
D41& 0.623 & 0.629 & 0.623 & 0.620 & 0.623 & 0.621 & 0.995 &	0.990	& 0.997\\
D74& 0.854 &	0.855	& 0.855 & 0.851 &	0.848	& 0.849 & 0.996 &	0.992 &	0.993\\
heart-statlog & 0.822 &	0.822 &	0.822 & 0.822	& 0.822	& 0.822 & 1.00 & 1.00 & 1.00 \\
liver-disorders & 0.624 &	0.629 &	0.625 & 0.620	& 0.623	& 0.621 & 0.994	& 0.990 &	0.994\\
monks-problems-1 & 0.459 &	0.461	& 0.456 & 0.397 &	0.398 &	0.397 & 0.865	& 0.863	& 0.871\\
sick & 0.987 &	0.987	& 0.987 & 0.979	& 0.979 &	0.979 & 0.992 &	0.992	& 0.992\\
\hline
\end{tabular}
\vspace*{-0.5cm}
\end{table*}

To facilitate comparisons between the recommended and label methods, we introduce three metrics: $\mathcal{F}^1= \frac{Precision_1}{Precision_2}$, $\mathcal{F}^2= \frac{Recall_1}{Recall_2}$, $\mathcal{F}^3= \frac{F-measure_1}{F-measure_2}$.

The larger the value of these indicators, the more similar the effect of the two approaches are. If the two methods are consistent, all of these indicators equal to 1. As shown in Table~\ref{tab:per}, although the recommended technique is inconsistent with the label method, its performance still be relatively comparable to the label method. Except for the $monks-problems-1$ dataset, all of the other datasets had $\mathcal{F}^1$, $\mathcal{F}^2$, $\mathcal{F}^3$ values greater than 0.9. The method recommended in this module, especially in $heart-statlog$, has the same excellent performance as the label method. To summarize, this module is capable of recommending efficient and effective decision methods for different problems.

\subsubsection{Explainability}
\begin{figure}[h]
\vspace*{-0.2cm}
  \centering
  \includegraphics[width=\linewidth]{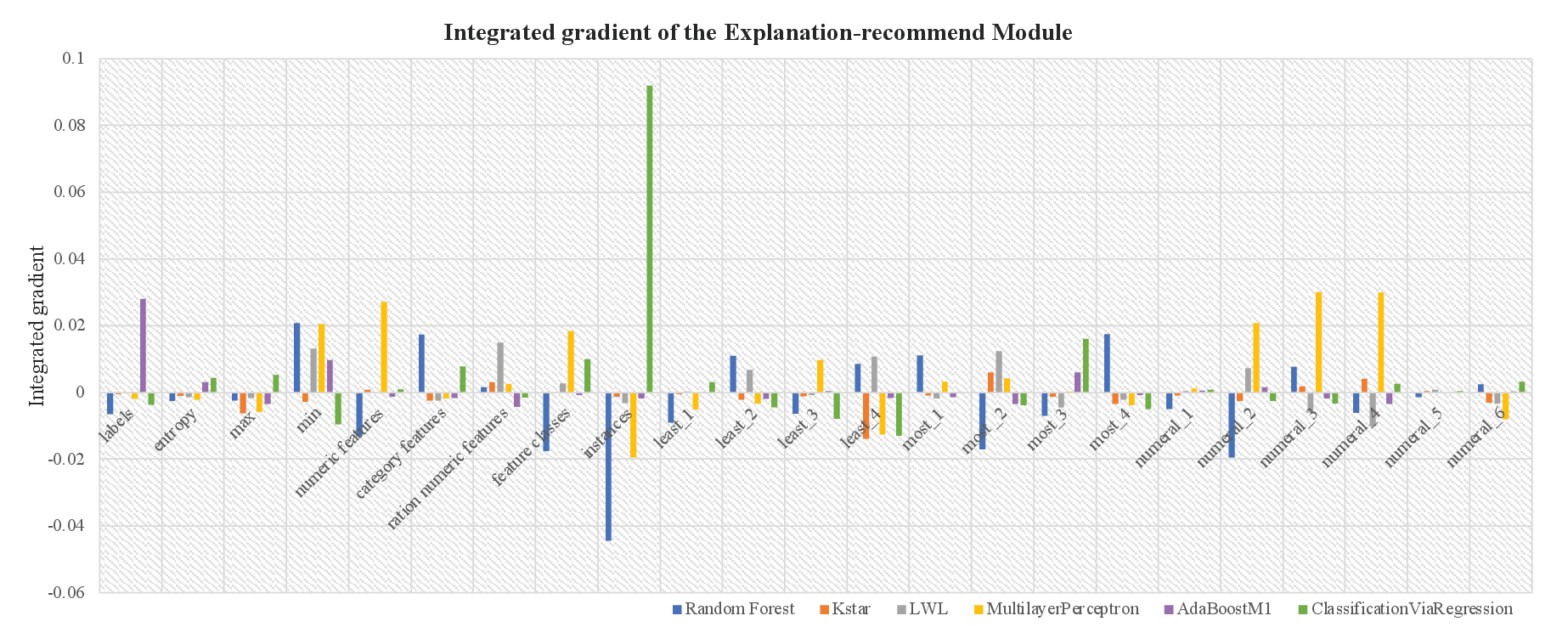}
   \caption{Integrated gradient of the meta-learning network }
   \label{gradients}
 \vspace*{-0.5cm}
 \end{figure}

For explainability, we test the recommendation network in terms of integrated gradients and makes an attempt to establish a link between some tough datasets and successful decision methods based on them. As illustrated in Figure \ref{gradients}, the adaboostM1 is an improved method of adaboost that is more adaptable to multi-class single-label tasks. The integrated gradient of this method is significantly larger than that of other methods on the feature of the number of target attribute classes, making it superior to other methods in multi-class tasks. The random forest is more capable than other approaches for classification problems involving a large number of category features, and our studies also demonstrate that the integrated gradient of this method is greater than that of other methods when the number of category attributes is considered.

Additionally, our experiments demonstrate that the method performs better as the minimum proportion of individual categories in the target attribute or the minimum proportion of individual categories in the category attribute with the most categories in the category attribute increases, but its performance may degrade as the category attribute with the most categories in the category attribute's information entropy or the maximum average value in the numeral increases. Besides, MLP differs from random forest in that it is better suited to datasets with a large number of numerical features, as evidenced by the fact that its integrated gradient is positive regardless of the maximum mean, minimum variance, or maximum variance in numerical attributes, and that its gradient value is significantly higher than that of other methods when the number of numerical attributes is taken into consideration. The ClassficationViaRegression method for classification by regression requires more instances to achieve a fit for a more accurate classification.

\subsection{Evaluation on feature influence computation}
The purpose of this part is to evaluate whether the feature influence scores proposed in this module more accurately capture the causal relationships in the domain.

\subsubsection{Experimental settings}
In this part of the evaluation, we consider two real-world datasets (Adult and German-Credit datasets). The Adult dataset contains information on 14 categories, including age, gender, occupation, education level, and so on. The objective is to forecast if an individual's annual income reaches 50k. The German-Credit covers 24 dimensions of personal, financial, and demographic information about the bank account holder. The objective is to categorize loan borrowers as having a good or bad credit risk. Additionally, we apply the causal graphs provided in the \cite{chiappa2019path} for both the Adult and German-Credit datasets.
\subsubsection{Baselines}
Lime \cite{ribeiro2016should}:This method allows visualization of feature importance scores and feature heatmaps to provide interpretation. SHAP \cite{lundberg2017unified}:an interpretation method based on cooperative game theory that provides an interpretation for black box models by calculating the Shapley value.
\subsubsection{Feature Influence Computation}
\begin{figure}[htbp]
\centering
\begin{minipage}{4cm}
\centering
\includegraphics[width=4cm]{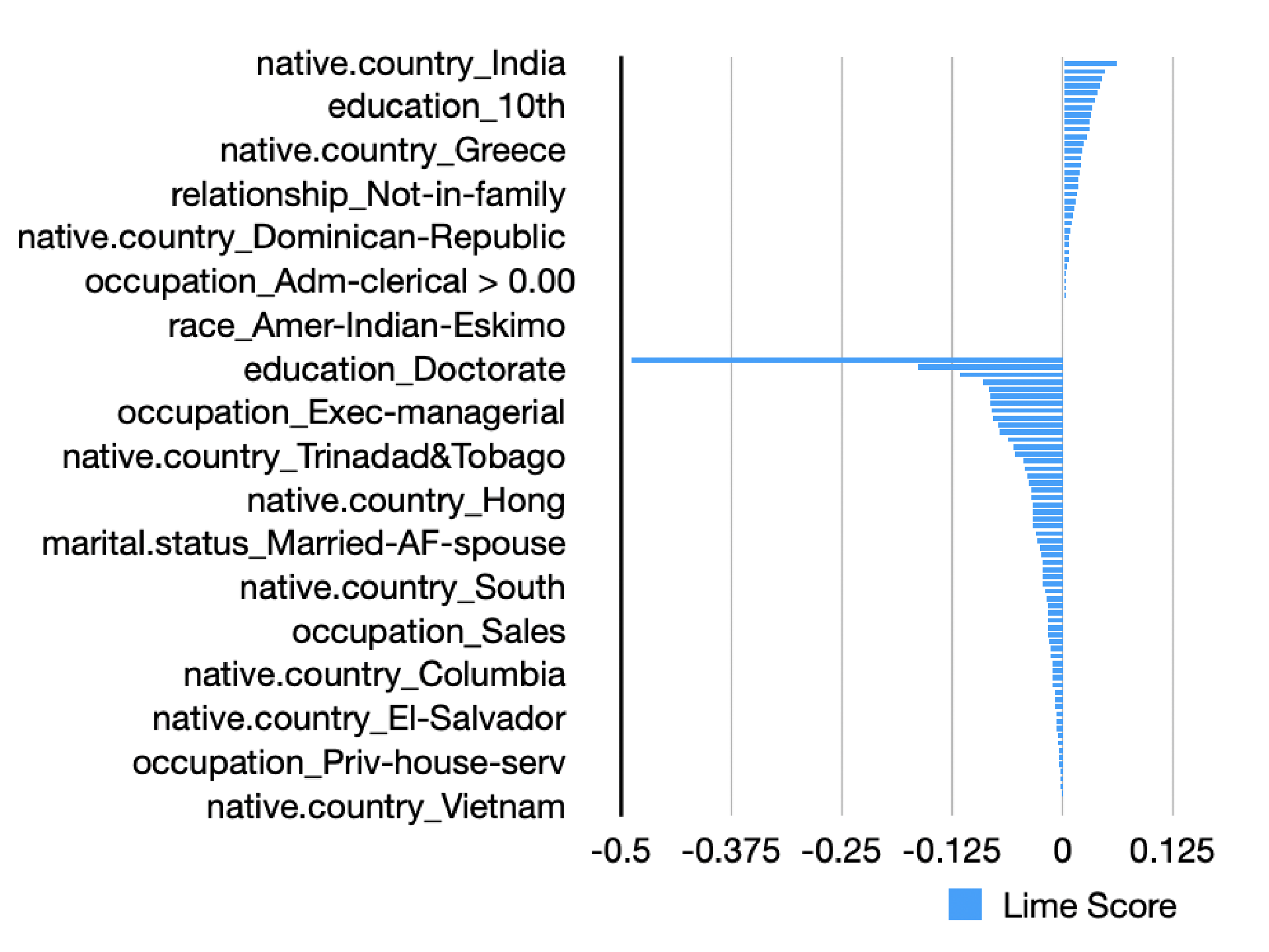}
\caption{Lime on Adult}
\label{arf}
\end{minipage}
\begin{minipage}{4cm}
\centering
\includegraphics[width=4cm]{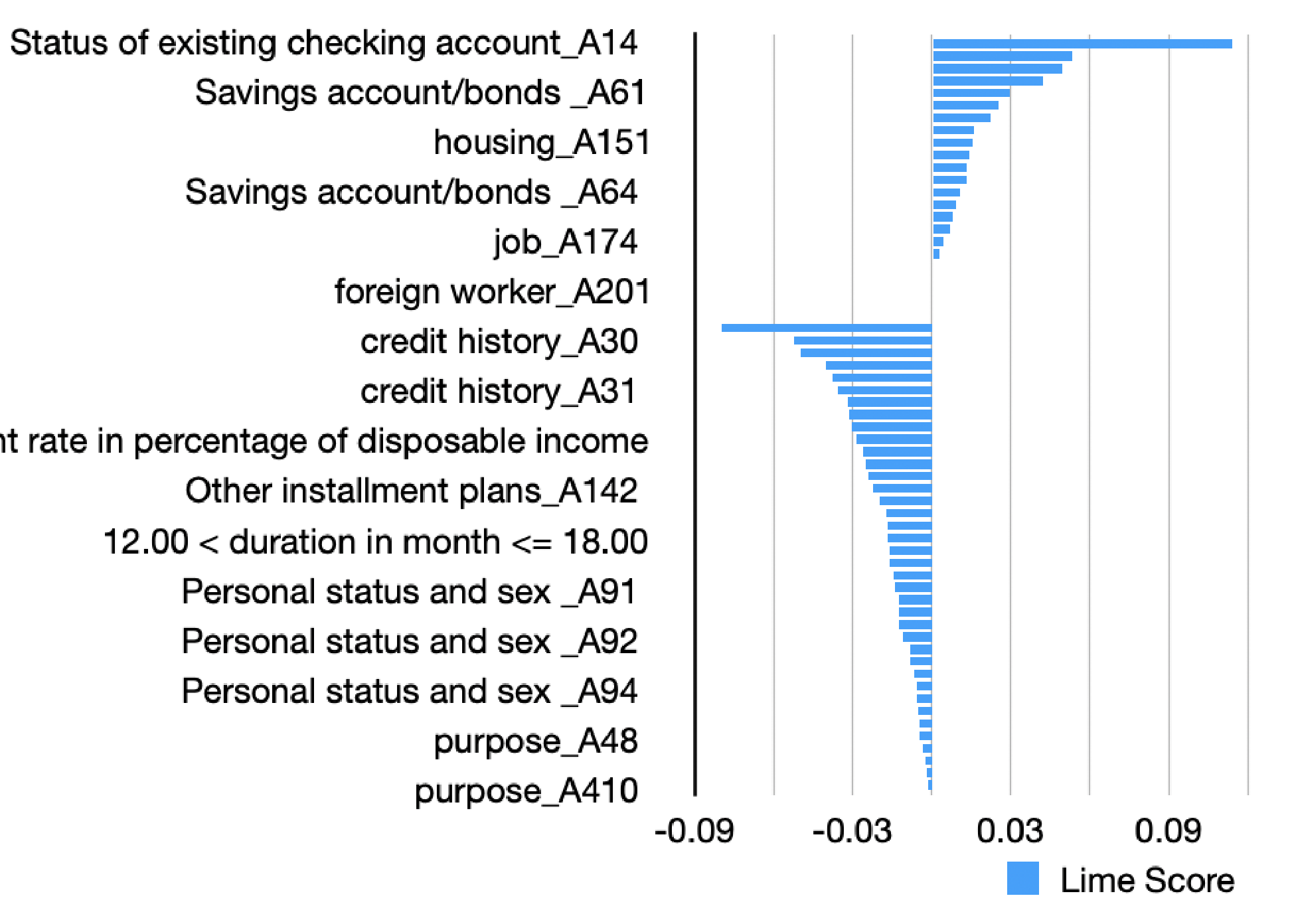}
\caption{Lime on German-Credit}
\label{grf}
\end{minipage}
\vspace*{-0.5cm}
\end{figure}

\begin{figure}[htbp]
\centering
\begin{minipage}{4cm}
\centering
\includegraphics[width=3cm]{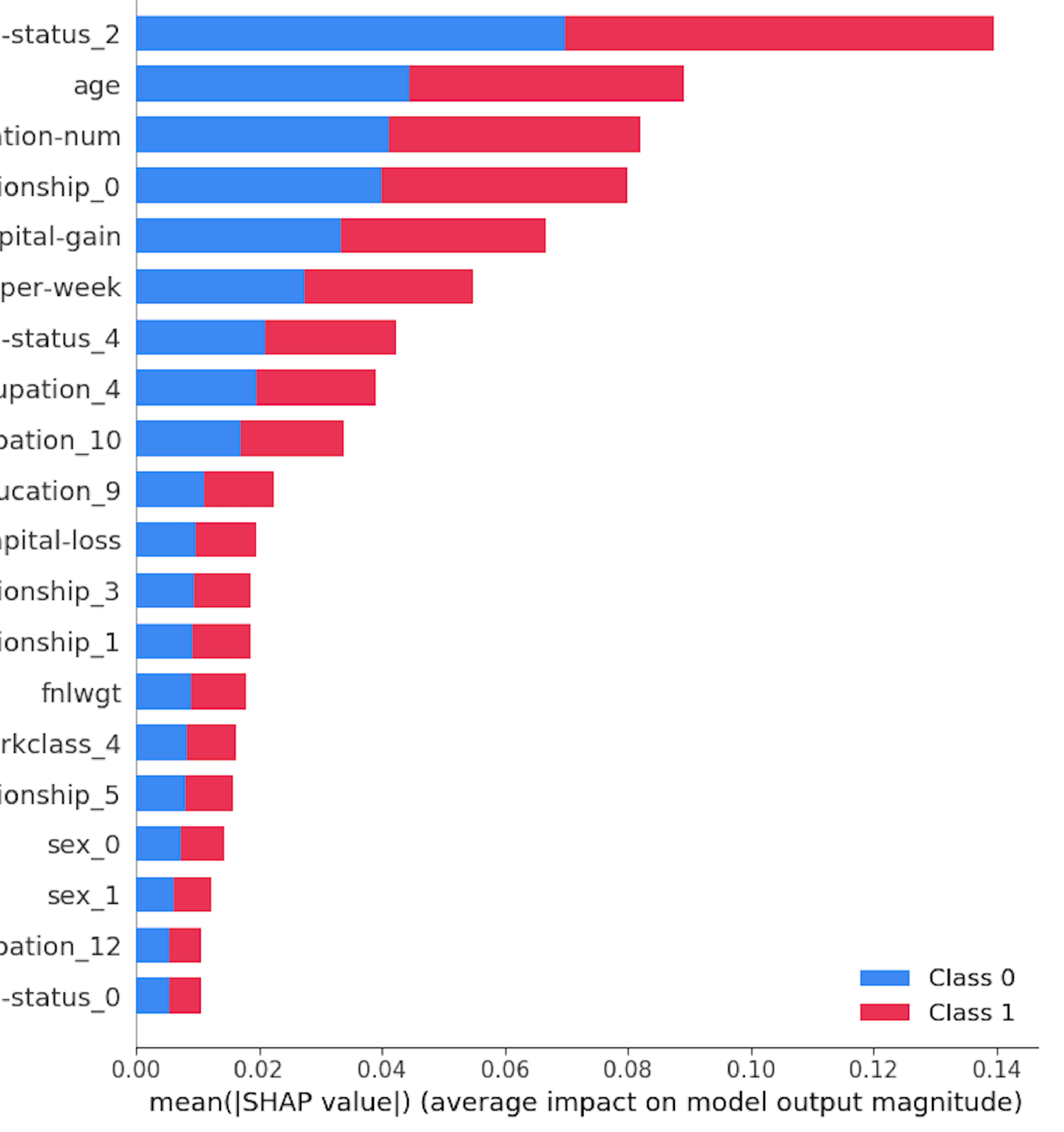}
\caption{SHAP on Adult}
\label{ash}
\end{minipage}
\begin{minipage}{4cm}
\centering
\includegraphics[width=3cm]{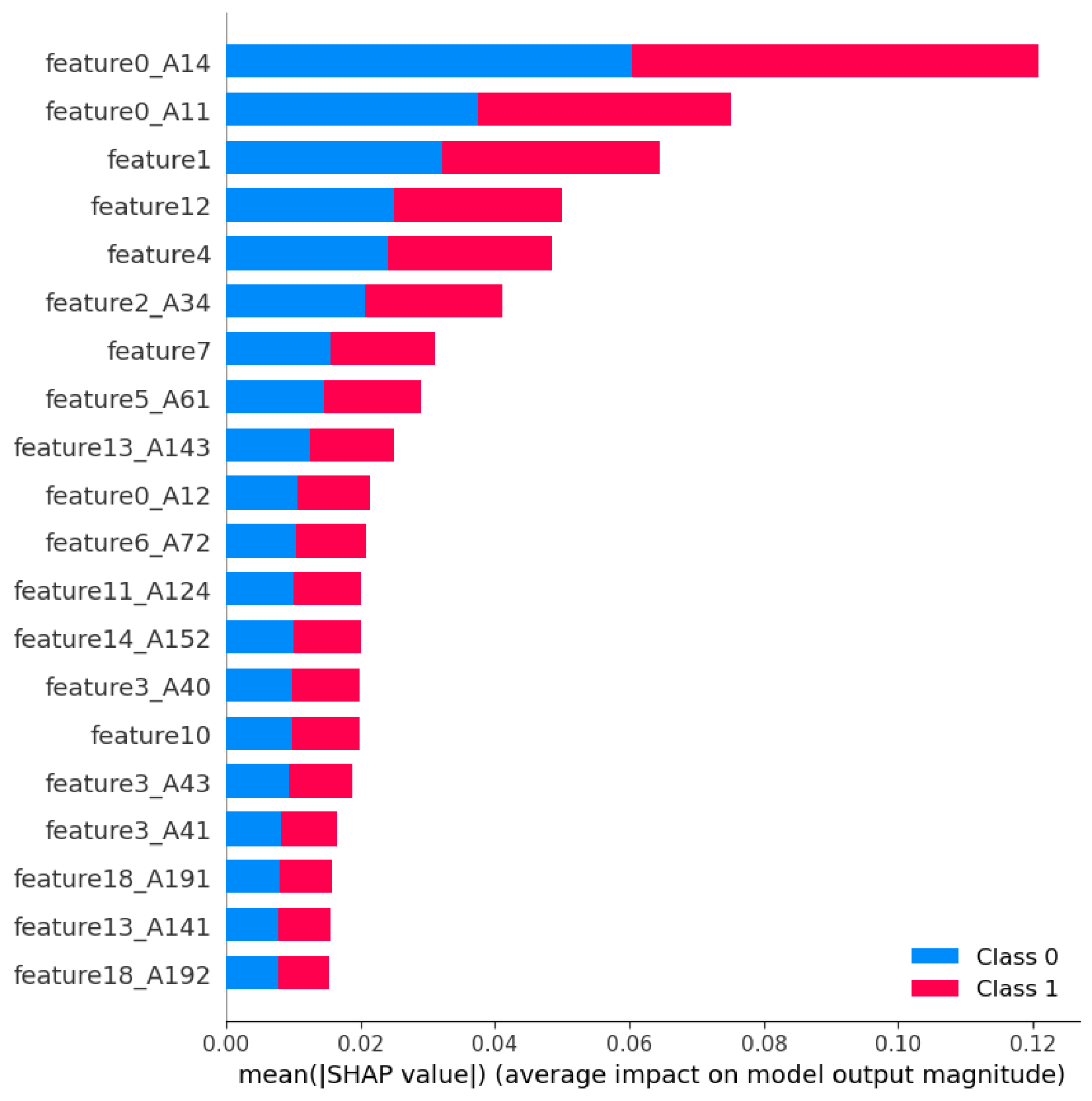}
\caption{SHAP on German-Credit}
\label{gsh}
\end{minipage}
\vspace*{-0.2cm}
\end{figure}

\begin{figure}[htbp]
\centering
\begin{minipage}{4cm}
\centering
\includegraphics[width=4cm]{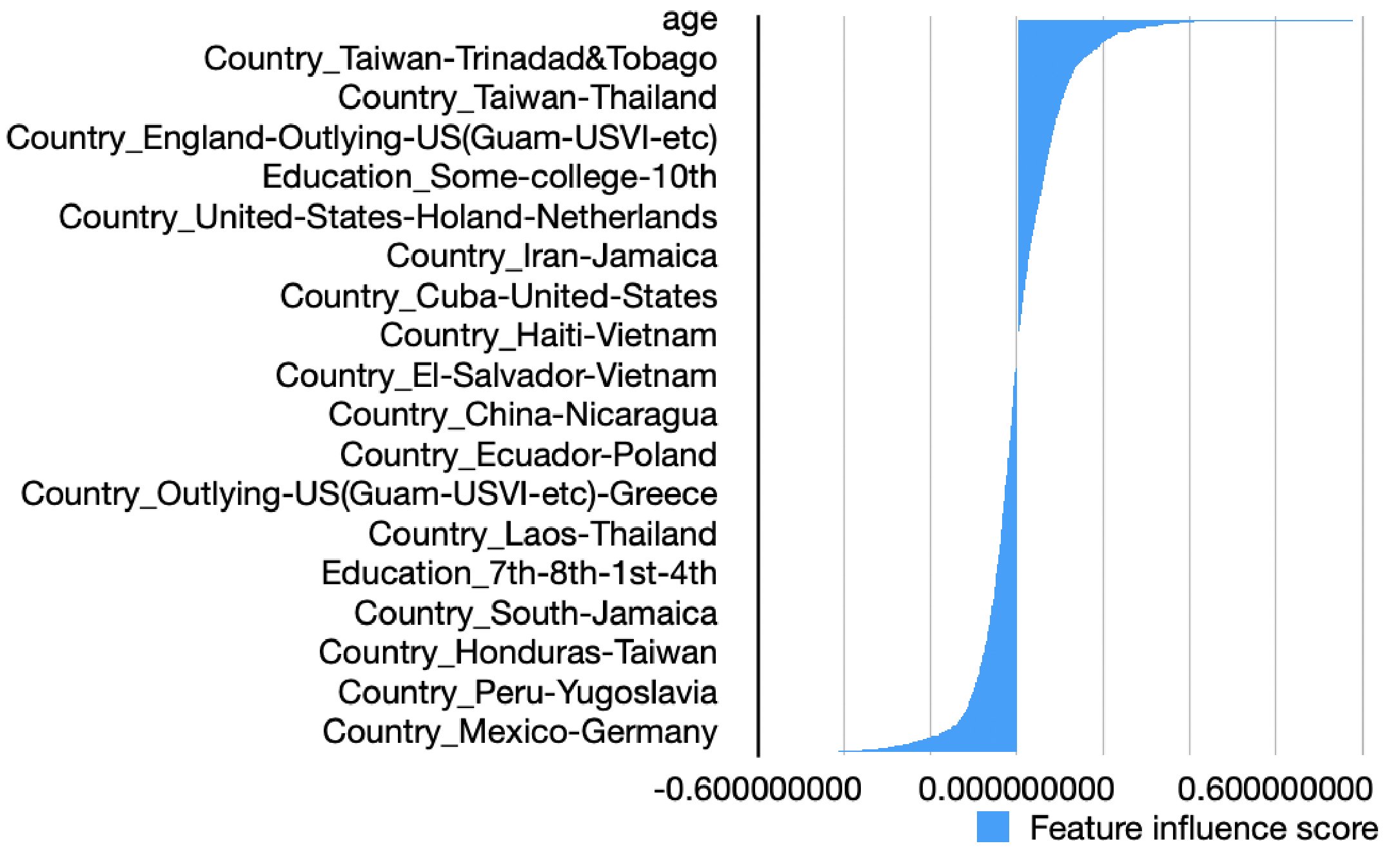}
\caption{Ours on Adult}
\label{ao}
\end{minipage}
\begin{minipage}{4cm}
\centering
\includegraphics[width=4cm]{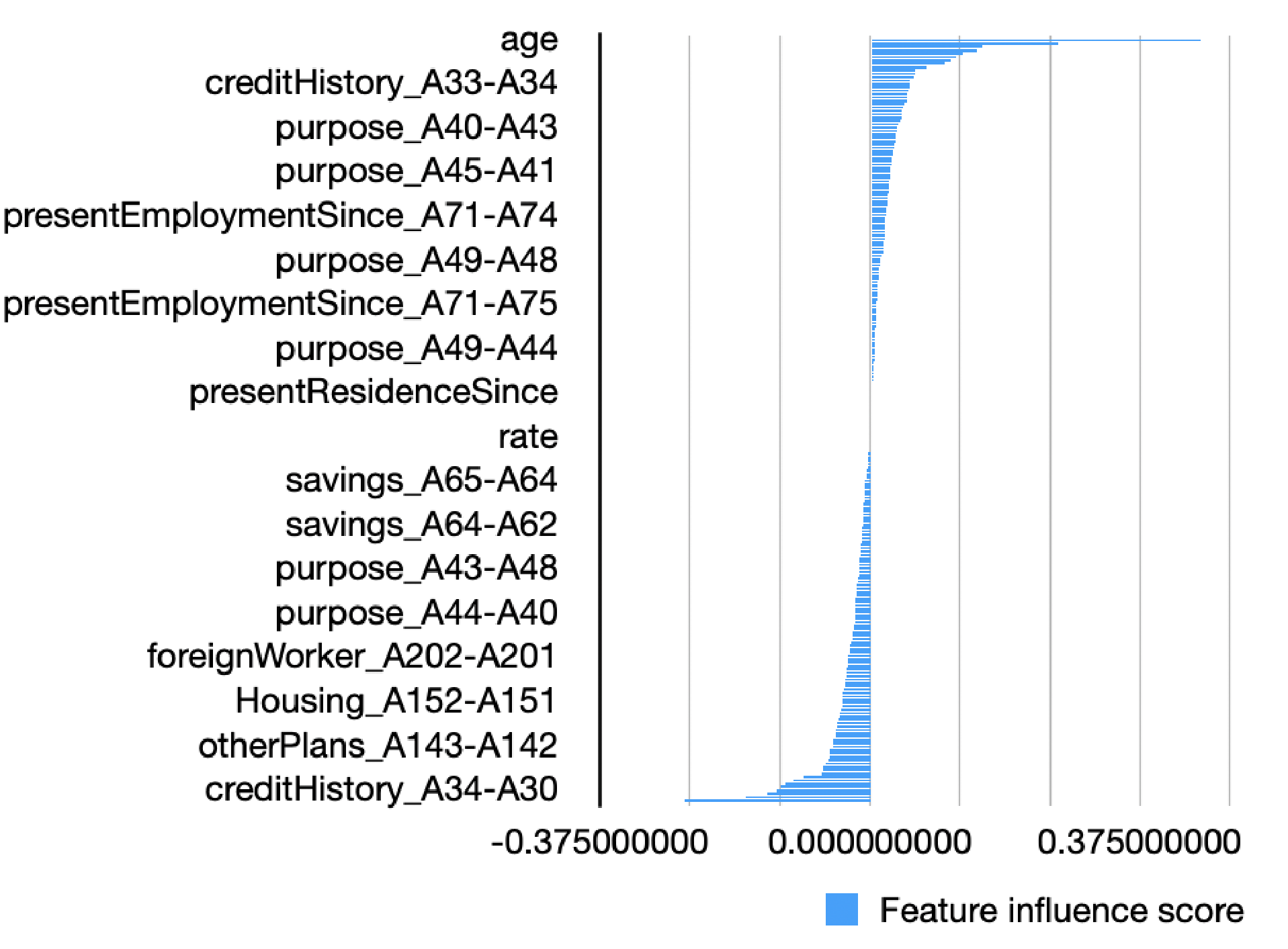}
\caption{Ours on German-Credit}
\label{go}
\end{minipage}
\vspace*{-0.3cm}
\end{figure}

Lime, SHAP, and our suggested technique all performed well on the two datasets, as illustrated in Figure \ref{arf}-\ref{go}. In comparison to the Lime and SHAP, our method calculates the impact scores of category features independently for each change in categorical characteristics, allowing us to more correctly capture the effect of specific feature values on the model. And our findings highlight causation, such as how age is typically associated with high income and how having housing or a growing credit history is likely to result in favorable outcomes for persons for whom the credit algorithm returns negative results.

\vspace*{-0.3cm}

\vspace*{-0.3cm}

\vspace*{-0.3cm}
\subsection{Evaluation on counterfactual generation}
In this subsection, we generate instance-specific counterfactuals based on the feature influence scores proposed in this paper, and adequately compare and evaluate them with existing popular counterfactual baseline methods in various aspects such as capability, efficiency and quality of counterfactual generation.

\subsubsection{Experimental settings}
We continue to use two real-world datasets, Adult and German-Credit, and the metrics chosen are centered on both reliability and efficiency.

\subsubsection{Baselines}
CEM \cite{dhurandhar2018explanations}: For the contrast counterfactual, the proximal gradient descent approach is employed. This is a very representative class of techniques, and as a result, we have chosen it as the comparison model for the baseline comparison.

Proto \cite{looveren2021interpretable}: To find interpretable counterfactual interpretations of classifier predictions, a method known as prototyping is used. The prototyping method uses class prototypes that have been obtained through class-specific K-d trees to speed up the search for counterfactual instances while also eliminating computational bottlenecks caused by the numerical gradient evaluation of black box models.

\subsubsection{Generated Counterfactuals}
\begin{figure}[htbp]
\centering
\begin{minipage}{4cm}
\centering
\includegraphics[width=4cm]{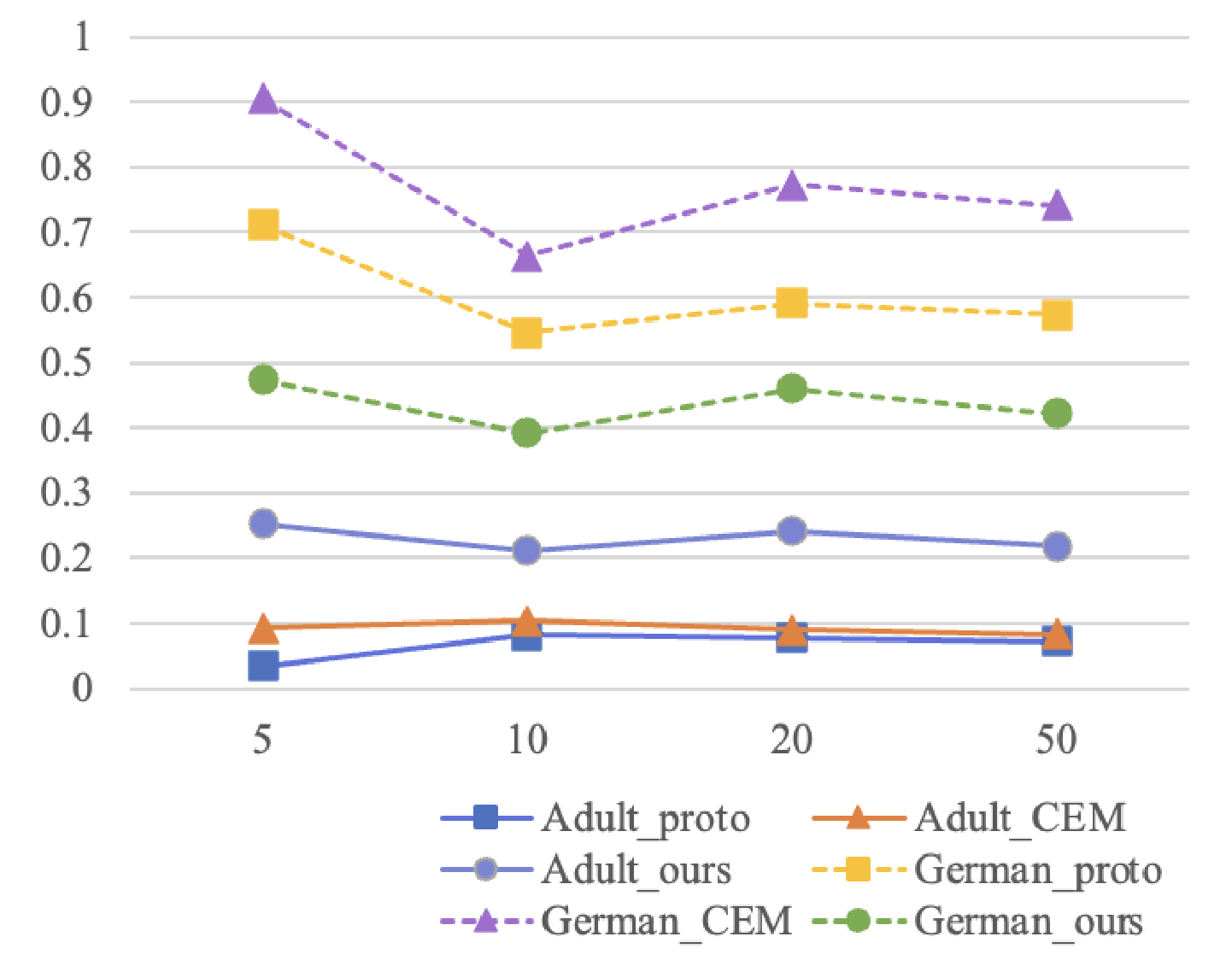}
\caption{Average distance}
\label{distance}
\end{minipage}
\begin{minipage}{4cm}
\centering
\includegraphics[width=4cm]{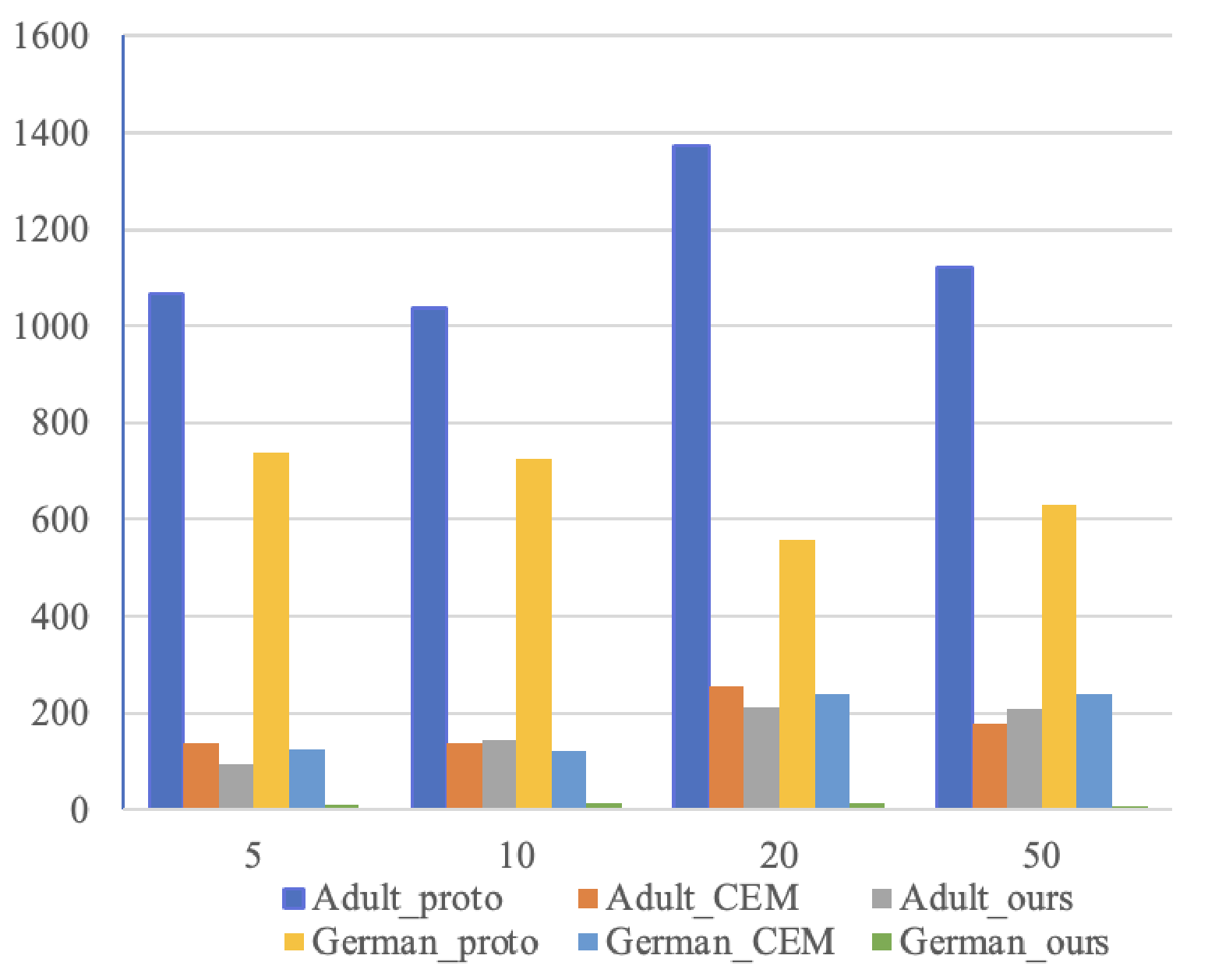}
\caption{Average time cost }
\label{timecost}
\end{minipage}
\vspace*{-0.5cm}
\end{figure}

In the specific experiments described in this study, 5, 10, 20, and 50 examples are randomly chosen for counterfactual generation. As illustrated in Figure~\ref{distance}, our method generates counterfactuals that are slightly more distant from the original instances than the CEM and Proto methods, since our method considers the causal relationship between features and is capable of dynamically updating the latent factors that are causally affected by the feature while it is modified. By contrast, the CEM and Proto methods disregard such causality, and in the pursuit of a greater distance from the original instance, these methods sacrifice the truthfulness of the generated counterfactuals, resulting in some implausible counterfactuals, such as the number of people whose census takers believe that the observation has been modified to a negative number, and so on.

As far as efficacy is concerned,shown in Figure \ref{timecost}, our strategy outperforms Proto by a wide margin. tThis is especially true on the larger Adult dataset, where the difference is more visible. Despite the fact that the generation efficiency of CEM on the Adult dataset is comparable to that of our method, on the smaller German-Credit dataset, CEM still requires a time investment comparable to that of it on the Adult dataset, whereas our method is able to generate counterfactuals on the German-credit dataset at a significantly lower time expenditure.
\vspace*{-0.3cm}

\section{conclusions}
This paper argues for the need for meta-learning explainability research, and divides meta-learning explainability into two aspects, explainability of the meta-learning process and explainability of meta-learning outputs in specific problems. This paper proposes and implements the FIND meta-learning explainability framework, in which model recommendation is used to achieve the meta explainability. And the recommendation explainability are achieved by
feature influence computation and counterfactual generation from the perspective of feature attribution and counterfactual, respectively. Future extensions of the work may include explainability in more challenging meta-learning domains, such as meta-learning of time series; and deeper counterfactual exploration incorporating causality.

\bibliographystyle{ACM-Reference-Format}
\bibliography{sample-base}

\appendix

\end{document}